\UseRawInputEncoding
\documentclass{article}

\usepackage{microtype}
\usepackage{graphicx}
\usepackage{subfigure}
\usepackage{booktabs} % for professional tables
\usepackage{multirow}
\usepackage{pifont}

\usepackage{hyperref}

\usepackage[accepted]{icml2025_modified}

\usepackage{amsmath}
\usepackage{amssymb}
\usepackage{mathtools}
\usepackage{amsthm}

\usepackage[capitalize,noabbrev]{cleveref}

\theoremstyle{plain}

\theoremstyle{definition}

\theoremstyle{remark}

\usepackage[textsize=tiny]{todonotes}

\icmltitlerunning{D2C: Unlocking the Potential of Continuous Autoregressive Image Generation with Discrete Tokens}

\begin{document}

\twocolumn[
\icmltitle{D2C: Unlocking the Potential of Continuous Autoregressive Image Generation with Discrete Tokens}
\icmlsetsymbol{internship}{*}
\icmlsetsymbol{correspondingauthor}{\textdagger}

\newcounter{@sch}\setcounter{@sch}{1}
\newcounter{@comp}\setcounter{@comp}{2}

\expandafter\gdef\csname the@sch\endcsname{1}
\expandafter\gdef\csname the@comp\endcsname{2}

\begin{icmlauthorlist}
\icmlauthor{Panpan Wang}{sch,internship}
\icmlauthor{Liqiang Niu}{comp}
\icmlauthor{Fandong Meng}{comp}
\icmlauthor{Jinan Xu}{sch,correspondingauthor}
\icmlauthor{Yufeng Chen}{sch}
\icmlauthor{Jie Zhou}{comp}
%\icmlauthor{}{sch}
\end{icmlauthorlist}
\icmlaffiliation{sch}{Beijing Key Laboratory of Traffic Data Mining and Embodied Intelligence, Beijing Jiaotong University, Beijing, China }
\icmlaffiliation{comp}{Pattern Recognition Center, WeChat AI, Tencent Inc, China}
\icmlcorrespondingauthor{Panpan Wang}{panpwang@bjtu.edu.cn}
\icmlcorrespondingauthor{Jinan Xu}{jaxu@bjtu.edu.cn}

\begin{center}
\textsuperscript{1}Beijing Key Laboratory of Traffic Data Mining and Embodied Intelligence, \\Beijing Jiaotong University, Beijing, China \\
\textsuperscript{2}Pattern Recognition Center, WeChat AI, Tencent Inc, China \\
\texttt{\{panpwang,jaxu,chenyf\}@bjtu.edu.cn} \\
\texttt{\{poetniu,fandongmeng,withtomzhou\}@tencent.com}
\end{center}

% \icmlkeywords{Machine Learning, ICML}

\vskip 0.2in
]

% this must go after the closing bracket ] following \twocolumn[ ...

% This command actually creates the footnote in the first column
% listing the affiliations and the copyright notice.
% The command takes one argument, which is text to display at the start of the footnote.
% The \icmlEqualContribution command is standard text for equal contribution.
% Remove it (just {}) if you do not need this facility.

\printAffiliationsAndNotice{\internship\\\indent\correspondingauthor}  % leave blank if no need to mention equal contribution
% \printAffiliationsAndNotice{\icmlEqualContribution} % otherwise use the standard text.

\begin{abstract}
In the domain of image generation, latent-based generative models occupy a dominant status; however, these models rely heavily on image tokenizer. To meet modeling requirements, autoregressive models possessing the characteristics of scalability and flexibility embrace a discrete-valued tokenizer, but face the challenge of poor image generation quality. In contrast, diffusion models take advantage of the continuous-valued tokenizer to achieve better generation quality but are subject to low efficiency and complexity. The existing hybrid models are mainly to compensate for information loss and simplify the diffusion learning process. The potential of merging discrete-valued and continuous-valued tokens in the field of image generation has not yet been explored. In this paper, we propose \textbf{D2C}, a novel two-stage method to enhance model generation capacity. In the first stage, the discrete-valued tokens representing coarse-grained image features are sampled by employing a small discrete-valued generator. Then in the second stage, the continuous-valued tokens representing fine-grained image features are learned conditioned on the discrete token sequence. In addition, we design two kinds of fusion modules for seamless interaction. On the \textit{ImageNet-256} benchmark, extensive experiment results validate that our model achieves superior performance compared with several continuous-valued and discrete-valued generative models on the class-conditional image generation tasks.
\end{abstract}

\section{Introduction}
During these years, discrete-valued generative models have made remarkable progress in the realm of image generation \cite{yu2022scaling, xie2024show, wu2024vila, liu2024lumina, yu2024randomized}. The discrete image tokenizer that converts encoded features of the image into discrete-valued tokens through vector quantization \cite{van2017neural, esser2021taming}, is one of the particularly important components of autoregressive and masked image generation models. Afterwards, discrete image tokens are applied to various image understanding and generation tasks with the same paradigm as language tokens. Autoregressive model is one of the most popular solutions because of outstanding achievements in the field of linguistics and their own versatility and scalability. Autoregressive image generation models \cite{yu2021vector, sun2024autoregressive, ma2024star, zhang2024var, yu2024image, luo2024open, yu2024randomized} following the next token prediction paradigm have shown great performance in text-to-image generation and class-to-image generation tasks. In addition to this, non-autoregressive approaches \cite{chang2022maskgit, li2023mage, chang2023muse, yu2024image, chen2024maskmamba} employ masked image models to predict multiple randomly masked tokens with bidirectional attention, which conforms to the bidirectional correlations essence of visual signals and accelerates the sampling speed in model inference process.

Concurrently, another predominant methods are continuous-valued generative models, which employ a continuous tokenizer to encode and decode the image without vector quantization \cite{kingma2013auto}. Diffusion models \cite{song2020denoising, ho2020denoising} are the most common methods in continuous-valued generative models, which recover latent space of image from random Gaussian noise through progressively denoising, such as LDM with the U-Net block \cite{rombach2022high}, DiT with the transformer block \cite{peebles2023scalable}. In addition, methods about merging autoregressive model and diffusion model \cite{tschannen2023givt, fan2024fluid, zhou2024transfusion, zhao2024monoformer} are beginning to emerge. Although diffusion models achieve slightly better generation performance than discrete-valued generative models, the challenges of low computational efficiency and complicated structure are still unavoidable. Subsequently, MAR model \cite{li2024autoregressive} uses a small MLP network to iteratively denoise with autoregressive prior that achieves the trade-off between accuracy and speed in image generation.

However, it does not matter whether discrete-valued generative models or continuous-valued generative models, each still faces some challenges. Firstly, the discrete image tokenizer in discrete-valued generative models reveals poor reconstruction quality because quantizing continuous encoded image features to discrete-valued tokens results in information loss. Secondly, high computational complexity of continuous-valued generative models constrains training and inference speed while the complete model is applied to iteratively denoise on the entire image. Third, research into merging discrete-valued tokens with continuous-valued tokens has not yet been well developed.

To address these challenges, we introduce \textbf{D2C}, a simple and effective hybrid model that improves image generation quality through synergizing discrete-valued and continuous-valued tokens. Specifically, it is a two-stage autoregressive framework. In the first phase, we train a small discrete-valued autoregressive model to sample the discrete token sequence as the coarse-grained image features that can be decoded into a low-quality image. In the next phase, we optimize a hybrid autoregressive model to generate continuous-valued tokens as the fine-grained image features conditioned on class token and generated discrete-valued tokens. Then, the decoder of VAE model is employed to convert the generated continuous-valued tokens into a high-quality image as our model final output. For realizing seamless interaction between discrete-valued and continuous-valued tokens, we elaborately design two kinds of fusion modules, cross-attention module and q-former module. Currently, there are a couple of work, such as DisCo-Diff \cite{xu2024disco}, HART \cite{tang2024hart} that are a little similar to our idea. But being different from them, discrete-valued and continuous-valued tokens are generated by different pretrained image tokenizers in our method and the two can be decoded separately, each yielding an image. Besides, our model is a two-stage framework to achieve coarse-grained to fine-grained image generation.

To summarize, our contributions are:
\begin{itemize}
\item We propose a novel two-stage hybrid autoregressive framework, D2C, which enhances image generation quality with fusing the discrete-valued and continuous-valued tokens from the independently trained image tokenizers.

\item We design two kinds of fusion modules including cross-attention module and q-former module, to achieve seamless interaction between discrete-valued and continuous-valued tokens.

\item To verify our model effectiveness and versatility, we conduct experiments on the \textit{ImageNet-256} benchmark. Experimental results indicate that our model outperforms both previous continuous-valued and discrete-valued generative models on the class-to-image task. Our model is more faster than MAR on this task in inference.
\end{itemize}

\section{Related Work}
\textbf{Continuous-valued image generation models} mainly contain diffusion models and autoregressive models. Diffusion models \cite{ho2020denoising, rombach2022high, ho2022imagen, betker2023improving} are known as state-of-the-art methods due to high-quality visual generation, which generate the image through an iteratively denoising process on the entire image. At first, convolutional U-Net is the backbone architecture in latent diffusion models \cite{rombach2022high, podell2023sdxl}. To improve scalability and efficiency, the architecture of transformer from \citet{peebles2023scalable, bao2023all, chen2023pixart} is proposed to replace the U-Net. Currently, another research direction that synergizes the autoregressive model and the continuous-valued token is gradually rising. GIVT \cite{tschannen2023givt} employs continuous-valued tokens instead of discrete-valued tokens in an autoregressive sequence model. In addition, MAR \cite{li2024autoregressive} autoregressively generates continuous-valued tokens equipped with diffusion loss, while Fluid \cite{fan2024fluid} scales up this idea to the text-to-image generation task. Besides, Transfusion \cite{zhou2024transfusion} and MonoFormer \cite{zhao2024monoformer} achieve autoregressive prediction and diffusion denoising on only one transformer architecture. And DART \cite{gu2024dart} supports autoregressively denoising within a non-Markovian diffusion framework to take full advantage of the generation trajectory. However, most of these models are limited by less efficient and complex diffusion structure.

\textbf{Discrete-valued image generation models} are based on autoregressive models and masked models. Pioneering research on this kind of models \cite{van2016pixel, chen2020generative} focuses on the pixel level in image generation. Along with the emergence of VQVAE \cite{van2017neural} and VQGAN \cite{esser2021taming}, which quantize per patch of image to the nearest discrete token, discrete-valued models \cite{sun2024autoregressive, yu2024randomized} have become one of the mainstream methods in the domain of image generation. In order to improve the performance of image tokenizer, various models have been proposed \cite{yu2021vector, lee2022autoregressive, zheng2022movq, huang2023not, huang2023towards, cao2023efficient}. However, there is still a serious problem, codebook collapse. Then regularization \cite{razavi2019generating, zhang2023regularized}, codebook transfer \cite{zhang2024codebook}, codebook optimization \cite{zheng2023online, shi2024taming}, lookup-free quantization \cite{yu2023language} and codebook decomposition \cite{li2024imagefolder, bai2024factorized} are designed to address the codebook collapse issue and enhance image reconstruction quality. Moreover, \citet{chang2022maskgit, chang2023muse, li2023mage} utilize masked image models to improve sampling speed, while \citet{weber2024maskbit} operates directly on bit tokens to make full use of the rich semantic information. There were also efforts to insert arbitrary token orders into autoregressive models to adapt inherently bidirectional correlations of visual signals \cite{yu2024randomized, pang2024randar}. However, information loss caused by vector quantization still remains a challenge in discrete-valued image generation models, which could weaken the model generative capacity.

In view of the fact that continuous-valued and discrete-valued models have demonstrated excellent performance separately, hybrid models fusing continuous-valued and discrete-valued tokens may be a new perspective in image generation models. DisCo-Diff \cite{xu2024disco} has attempted to augment the continuous diffusion model with complementary discrete-valued tokens, proving the feasibility and effectiveness of the hybrid model in image generation tasks. In addition, Hart \cite{tang2024hart} makes up for information loss of discrete-valued tokens with continuous-valued tokens. On this basis, we propose a two-stage hybrid autoregressive model to further improve the image generation quality following from coarse-grained to fine-grained rule.

\section{Preliminaries}
\textbf{Autoregressive image generation models} following the next token prediction paradigm generate a token sequence from left to right and one by one. Given a sequence of tokens $x=[x_1, x_2, \dots, x_n]$, where $n$ is the length of the sequence. The autoregressive model is trained to model the probability distribution of next token depending on the past generated tokens:
\begin{equation}
p_\theta(x)=\prod_{i=1}^{n}p_\theta(x_i|x_1,x_2, \dots, x_{i-1})
\label{eq_1}
\end{equation}
where $\theta$ denotes the model parameter. The goal of autoregressive model is to optimize the $p_\theta(x_i|x_1,x_2, \dots, x_{i-1})$ over all image token sequences that are generated by a discrete image tokenizer \cite{esser2021taming, sun2024autoregressive, yu2024randomized}. In general, the autoregressive model consists of a stack of transformer blocks with causal attention, as shown in Eq.\ref{eq_1}. In order to satisfy the unidirectional characteristics of the autoregressive model, each image is converted into a 1-dimensional sequence in the raster scan order.

\textbf{Masked image generation models} following the masked tokens prediction paradigm generate a sequence in an out-of-order manner. Analogously, the image is firstly encoded and quantized by a discrete image tokenizer and then the image tokens are randomly selected as masked tokens \cite{chang2022maskgit, chang2023muse, weber2024maskbit}. The masked models are optimized through maximizing the likelihood of the masked tokens conditioned on the unmasked tokens. The difference with the autoregressive models is that the masked models could predict multiple tokens at each step with bidirectional attention.

\textbf{Diffusion loss with MLP} is a novel loss function for each token that is a continuous-valued vector \cite{li2024autoregressive}. Diffusion loss follows the same principle of denoising to model a probability distribution of $x$ conditioned on both $t$ and $z$ following the diffusion models \cite{ho2020denoising, rombach2022high, peebles2023scalable}, denoted as $p(x|z, t)$:
\begin{equation}
\mathcal{L}(z,x)=\mathbb{E}_{\varepsilon, t}\left[\left\|\varepsilon-\varepsilon_{\theta}(x_{t}\vert t, z)\right\|^{2}\right]
\label{eq_2}
\end{equation}
Specifically, $t$ is a time step of the noise schedule sampled from $\left\{1, \dots, T\right\}$. $x_t$ is the noise-corrupted vector sampled from $x_{t}=\sqrt{\bar{\alpha}_{t}} x_{0}+\sqrt{1-\bar{\alpha}_{t}}\varepsilon$, a Markovian process, where $\bar{\alpha}_{t}$ is the hyper-parameter determined by the noise schedule, $x_0$ is original image and $\varepsilon$ is the sampled Gaussian noise. A noise prediction network $\varepsilon_{\theta}$ is parameterized by $\theta$, which is a small MLP network for iteratively denoising. In addition, $z$ is a conditioning vector produced by a generative model. The denosing MLP is trained by minimizing the mean-squared error between the predicted noise and the ground truth Gaussian noise.

\begin{figure*}[th] 
\vskip 0.1in
\begin{center}
\centerline{\includegraphics[width=\textwidth]{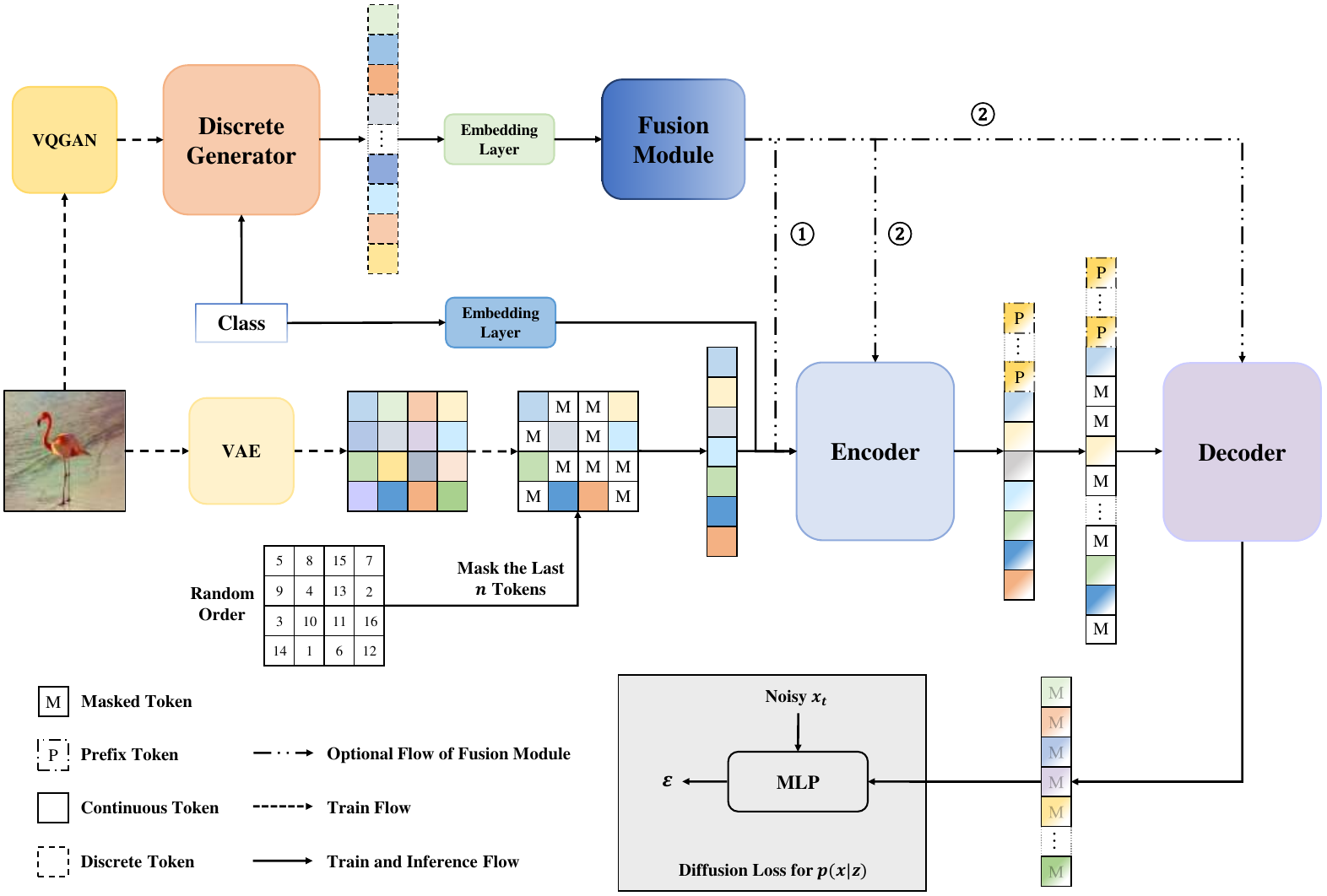}}
\caption{\textbf{Overview of our D2C framework.} It is a two-stage hybrid autoregressive model fusing discrete-valued and continuous-valued tokens. In this figure, we mainly illustrate the training and inference process of the second stage. The fusion module has two kinds of structures, cross-attention module and q-former module. The "Prefix token" is a class token or query token, and it only is a class token when the fusion module is the cross-attention module. When the fusion module is q-former module, the flow "\ding{172}" is chose, otherwise the flow "\ding{173}" is chose.}
\label{model_overview}
\end{center}
\vskip -0.1in
\end{figure*}

\section{Method}
Currently, enhancing image generative capacity with integration of discrete-valued and continuous-valued tokens is gaining prominence, and its effectiveness has been confirmed within the realm of diffusion models \cite{xu2024disco}. The existing method of fusing continuous-valued and discrete-valued tokens in the autoregressive models is mainly to compensate for information loss caused by the vector quantization \cite{tang2024hart}. But regretably, the maximum attainable performance of this method is limited to the performance exhibited by the continuous-valued generation model. To break the ceiling and further improve the generative capability, we propose D2C, a two-stage autoregressive hybrid model following a progressive generation manner that generates the continuous-valued tokens condition on discrete-valued tokens. The key point is how to gain meaningful discrete-valued tokens and how to connect discrete-valued tokens and continuous-valued tokens. 

\subsection{From Class to Discrete Tokens}
In the first stage, we aim to train a small discrete-valued autoregressive model to sample a discrete token sequence from a class token. These discrete-valued tokens of a image represent the coarse-grained image features, which will be as condition information in the next stage. Firstly, the input image $x\in \mathbb{R}^{H\times W\times 3}$ is encoded and then quantized into discrete-valued tokens $q\in \mathbb{R}^{h\times w}$ via the discrete image tokenizer, VQGAN \cite{esser2021taming}. Here, $r=W/w=H/h$ denotes the downsampling ratio of the discrete image tokenizer and then these tokens are reshape a 1-dimensional sequence $q=[q_1,\dots, q_i,\dots,q_{h\cdot w}]$ in raster scan order, where $h\cdot w$ is the length of the sequence and $q_i\in q$ is indices from the codebook of discrete image tokenizer. Afterwards, we follow the Eq.\ref{eq_3} to train a small Transformer-based \cite{vaswani2017attention} autoregressive image generation model.
\begin{equation}
p_\theta(q)=\prod_{i=1}^{h\cdot w}p_\theta(q_i|q_1, \dots, q_{i-1}, c)
\label{eq_3}
\end{equation}
where $q_{<i}$ denote the all preceding tokens and $c$ is the corresponding class token embedding, which is placed at the beginning of the sequence as a prefix. Following \citet{sun2024autoregressive}, the discrete-valued autoregressive model adopts similar architecture with Llama \cite{touvron2023llama}. And we separately apply RMSNorm \cite{zhang2019root} and SwiGLU \cite{shazeer2020glu} for normalization and activation functions. 2D RoPE \cite{su2024roformer} is used as the positional embeddings. Therefore we get a autoregressive class-to-image generation model.

\begin{figure}[t]
\vskip 0.1in
\begin{center}
\centerline{\includegraphics[width=\columnwidth]{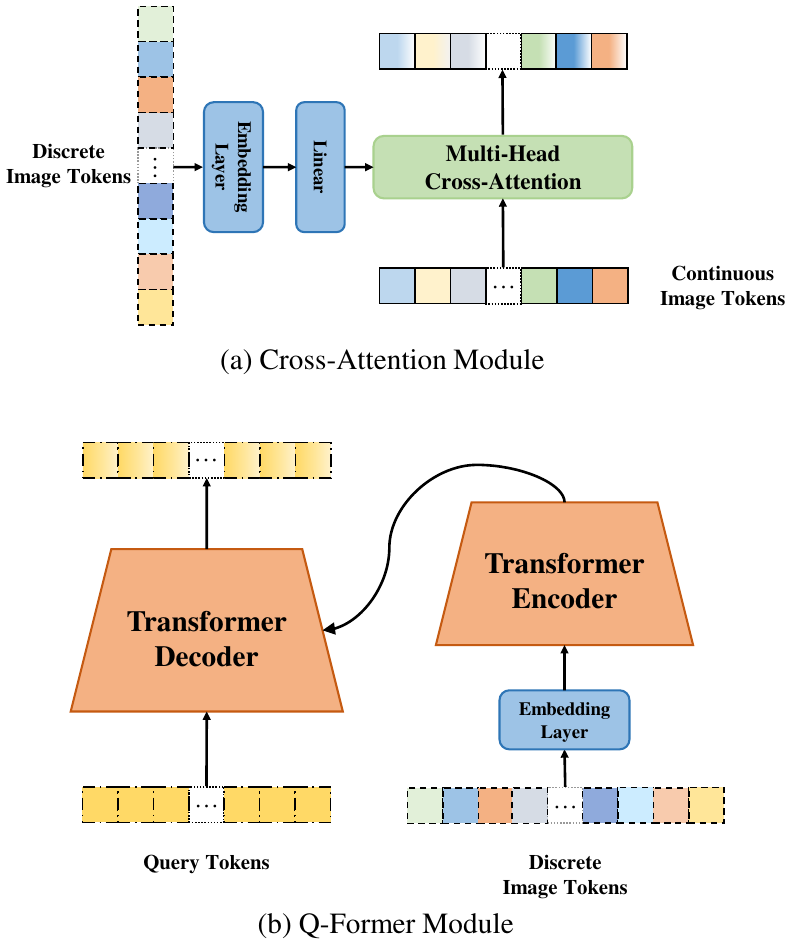}}
\caption{\textbf{Fusion Module.}}
\label{fusion_module}
\end{center}
\vskip -0.1in
\end{figure}

\subsection{From Discrete to Continuous Tokens}
In the second stage, the goal is to train the main body of our framework, a hybrid autoregressive model that generates the fine-grained image features, continuous-valued tokens, conditioned on the above discrete-valued tokens, as illustrated in Figure.\ref{model_overview}. We adopt the Masked Autoencoder (MAE) \cite{he2022masked} with bidirectional attention as our hybrid autoregressive model. In training of this stage, we use the above discrete-valued model to generate the next discrete-valued token conditioned on the class token and ground true tokens before this position that are produced by the discrete image tokenizer. For better interacting between discrete-valued and continuous-valued tokens, we insert a fusion module into the MAE. Specifically, we employ the continuous image tokenizer \cite{kingma2013auto} to encode the input image $x\in \mathbb{R}^{H\times W\times 3}$ as continuous-valued tokens $z\in \mathbb{R}^{h\times w\times d}$. The downsampling rate is same as the discrete image tokenizer. Then we randomly sample a masking rate $m_r$ from a truncated Gaussian distribution following \citet{li2023mage}. Uniformly, we convert the 2-dimensional continuous-valued image tokens into a 1-dimensional sequence $z_c=[z_{c,1},\dots, z_{c,i}, \dots, z_{c,h\cdot w}]$ and then mask out $m_r\cdot h\cdot w$ tokens which is replaced with a learnable mask token $[M]$. Besides, 64 identical class tokens are padded as the prefix of the input sequence for improving the stability and capacity of encoding \cite{li2024autoregressive}. In the encoder of MAE, the encoding
process is to primarily aggregate available and reliable information. In the decoder of MAE, the $[M]$ tokens gradually acquire the important information to generate the indispensable condition $z$. Afterwards, we utilize a lightweight MLP to iteratively denoise conditioned on $z$ and timestep $t$, while the training objective is to calculate matching scores utilizing mean-squared error, as shown in the Eq.\ref{eq_2}. For executing autoregressive generation, we introduce random orders \cite{li2024autoregressive}, where the input continuous-valued token sequence is shuffled at random. In training, we mask out the last $m_r\cdot h\cdot w$ tokens in the shuffled sequence. And we predict token of next position in the generated random order in inference. 

\subsection{Fusion Modules}
For connecting the two stages to achieve progressive generating, we propose two types of fusion module: cross-attention module and q-former module.

\textbf{Cross-Attention Module.} Our cross-attention module is shown in the Figure.\ref{fusion_module} (a), which is preceded by a normalization layer. In the MAE \cite{he2022masked}, the original block in encoder and decoder only contain self-attention and feedforward module. When the cross-attention module is selected as the fusion module, we add our cross-attention module between self-attention and feedforward module to carry out interaction. The cross-attention module is formulated as follows:
\begin{gather}
h_q=Linear_\theta(Embed_\theta(q)) \\[8pt]
Q = W_qh_c, \ K = W_kh_q, \ V = W_vh_q \\
CrossAttention(Q, K, V) = Softmax(\frac{QK^T}{\sqrt{d}}V)
\end{gather}

where $h_c\in \mathbb{R}^{hw\times d}$ and $h_q\in \mathbb{R}^{hw\times d}$ severally denote the hidden states of continuous-valued tokens and discrete-valued tokens. $W_q\in\mathbb{R}^{d\times d}$, $W_k\in\mathbb{R}^{d\times d}$, $W_v\in\mathbb{R}^{d\times d}$ are the learnable projection matrices, respectively.

\begin{table*}[th]
\caption{\textbf{The performance of our model on ImageNet $256\times256$ benchmark.} "Token" refers to token type used by the generation model where "discrete" and "continuous" denote the token generated by discrete tokenizer (eg. VQGAN) and continuous tokenizer (eg. VAE), respectively. "both" contains the above two type of tokens. "Type" is the type of generation model including diffusion model (Diff.), masked model (Mask.), autoregressive model (AR) and visual autoregressive model (VAR). "\#Step" is the number of inference step. In this table, the parameters of our model contain discrete-valued generation model (111M) and continuous-valued generation model.}
\label{table_1}
\vskip 0.1in
\begin{center}
\begin{footnotesize}
\setlength{\tabcolsep}{2.5pt}
\renewcommand{\arraystretch}{1}
\begin{tabular}{cc|ccc|cccc|cccc}
\toprule
&  &  &  &  & \multicolumn{4}{c|}{with CFG} & \multicolumn{4}{c}{without CFG} \\
\textbf{Token} & \textbf{Type} & \textbf{Model} & \textbf{\#Params} & \textbf{\#Steps} & \textbf{FID\textdownarrow} & \textbf{IS\textuparrow} & \textbf{Pre.\textuparrow} & \textbf{Rec.\textuparrow} & \textbf{FID\textdownarrow} & \textbf{IS\textuparrow} & \textbf{Pre.\textuparrow} & \textbf{Rec.\textuparrow} \\
\midrule
\multirow{5}{*}{discrete} & Diff. & LDM-8 \cite{rombach2022high} & 258M & 200 & 7.76 & 209.5 & 0.74 & 0.35 & 15.51 & 79.0 & 0.65 & 0.63 \\
\cmidrule(lr){2-13}
& Mask. & MaskGIT \cite{chang2022maskgit} & 227M & 8 & 6.18 & 1.821 & 0.80 & 0.51 & - & - & - & \\
 % & \multirow{4}{*}{Mask.} & MaskGIT \cite{chang2022maskgit} & 227M & 8 & 6.18 & 1.821 & 0.80 & 0.51 & - & - & - & \\
 % &  & MAGVIT-v2 \cite{yu2023language} & 307M & 64 & 1.78 & 319.4 & - & - & 3.65 & 200.5 & - & - \\
 % &  & TiTok-S-128 \cite{yu2024image} & 287M & 64 & 1.97 & 281.8 & - & - & 4.44 & 168.2 & - & - \\
 % &  & MaskBit \cite{weber2024maskbit} & 305M & 256 & 1.52 & 328.6 & - & - & - & - & - & - \\
\cmidrule(lr){2-13}
 & \multirow{3}{*}{AR} & VQGAN \cite{esser2021taming} & 227M & 256 & 18.65 & 80.4 & 0.78 & 0.26 & - & - & - & \\
 &  & LlamaGen-B \cite{sun2024autoregressive} & 111M & 256 & 5.46 & 193.6 & 0.84 & 0.46 & 26.26 & 48.1 & 0.59 & 0.62 \\
 &  & LlamaGen-XL \cite{sun2024autoregressive} & 775M & 256 & 3.39 & 227.1 & 0.81 & 0.54 & 19.42 & 66.2 & 0.61 & \textbf{0.67} \\
 % &  & RAR-B \cite{yu2024randomized} & 261M & 256 & 1.95 & 290.5 & 0.82 & 0.58 & - & - & - & - \\
 % &  & RAR-L \cite{yu2024randomized} & 461M & 256 & 1.70 & 299.5 & 0.81 & 0.60 & - & - & - & - \\
\midrule
\multirow{5}{*}{continuous} & \multirow{2}{*}{Diff.} & LDM-4 \cite{rombach2022high} & 400M & 250 & 3.60 & 247.7 & \textbf{0.87} & 0.48 & 10.56 & 103.5 & 0.71 & 0.62 \\
 &  & DiT-XL/2 \cite{peebles2023scalable} & 675M & 250 & 2.27 & 278.2 & 0.83 & 0.57 & 9.62 & 121.5 & 0.67 & \textbf{0.67} \\
\cmidrule(lr){2-13}
 & \multirow{3}{*}{AR} & GIVT \cite{tschannen2023givt} & 304M & 256 & 3.35 & - & 0.84 & 0.53 & 5.67 & - & 0.75 & 0.59 \\
 &  & MAR-B \cite{li2024autoregressive} & 208M & 256 & 2.31 & 281.7 & 0.82 & 0.57 & 3.48 & 192.4 & 0.78 & 0.58 \\
 &  & MAR-L \cite{li2024autoregressive} & 479M & 256 & 1.78 & 296.0 & 0.81 & 0.60 & \textbf{2.60} & 221.4 & 0.79 & 0.60 \\
\midrule
\multirow{3}{*}{both} & VAR & HART-$d$20 \cite{tang2024hart} & 649M & 10 & 2.39 & \textbf{316.4} & - & - & - & - & - & - \\
\cmidrule(lr){2-13}
 & \multirow{4}{*}{AR} & D2C-B with Cross-Attetion Module & 389M & 64 & 2.25 & 266.8 & 0.77 & 0.63 & 3.58 & 227.9 & 0.82 & 0.52 \\
 \cmidrule(lr){3-13}
 &  & \multirow{2}{*}{D2C-B with Q-Former Module} & \multirow{2}{*}{362M} & 64 & 2.09 & 265.2 & 0.77 & 0.63 & 3.39 & 234.3 & 0.83 & 0.53 \\
 &  &  &  & 256 & 2.06 & 266.7 & 0.77 & 0.62 & 3.35 & 235.0 & 0.83 & 0.53 \\
 \cmidrule(lr){3-13}
  &  & \multirow{2}{*}{D2C-L with Q-Former Module} & \multirow{2}{*}{633M} & 64 & 1.73 & 285.7 & 0.77 & \textbf{0.64} & 3.14 & 264.5 & 0.84 & 0.53 \\
   &  &  &  & 192 & \textbf{1.71} & 285.1 & 0.78 & \textbf{0.64} & 3.20 & \textbf{269.3} & \textbf{0.85} & 0.53 \\
\bottomrule
\end{tabular}
\end{footnotesize}
\end{center}
\vskip -0.1in
\end{table*}

\textbf{Q-Former Module.} As shown in Figure.\ref{fusion_module} (b), the q-former module is a encoder-decoder architecture based on Transformer \cite{vaswani2017attention}. The input of q-former module include the features of discrete tokens $h_q\in \mathbb{R}^{hw\times d}$ and $L$ learnable query embeddings $e_{query}\in \mathbb{R}^{L\times d_{query}}$, where $L$ is the number of query embeddings and $d_{query}$ is the dim of query embedding. The query embeddings are introduced to refine coarse-grained image features from the sequences of $hw$ discrete-valued tokens generated in the first stage. This is similar with the queries introduced in BLIP-2 \cite{li2023blip}, GILL \cite{koh2024generating} and Minigpt-5 \cite{zheng2023minigpt} for extracting image features. Afterward, the output features $h_q\in \mathbb{R}^{L\times d}$ are concatenated with the class tokens and continuous-valued tokens. In the subsequent MAE model, the discrete-valued and continuous-valued tokens engage in self-attention interaction.
\begin{equation}
h_{query}=QFormer_\theta(e_{query}, Embed_\theta(q))
\end{equation}

\subsection{Discussions}
In the HART model \cite{tang2024hart}, the discrete-valued and continuous-valued tokens are generated by only a redesigned hybrid image tokenizer. The MLP in HART is employed for iteratively denoising to predict the residual tokens that offsets the information loss because of vector quantization. The performance of a continuous-valued generative model is its upper limit. Besides, in the Disco-Diff model \cite{xu2024disco}, a few complementary discrete-valued tokens are inferred through an encoder in training. The complementary discrete-valued tokens capture the global appearance patterns, such as style and color, but do not represent a specific and entire image. The motivation of Disco-Diff is to simplify the learning process of continuous-valued diffusion models. In our method, the discrete-valued and continuous-valued tokens are introduced by discrete and continuous image tokenizers, respectively. And we design a unique two-stage framework to realize progressive image generation. Although our diffusion MLP is same as MAR \cite{li2024autoregressive}, we fuse the two types of tokens with the fusion module to learn a more strong denoising condition.

\begin{figure*}[ht]
\vskip 0.1in
\begin{center}
\centerline{\includegraphics[width=\textwidth]{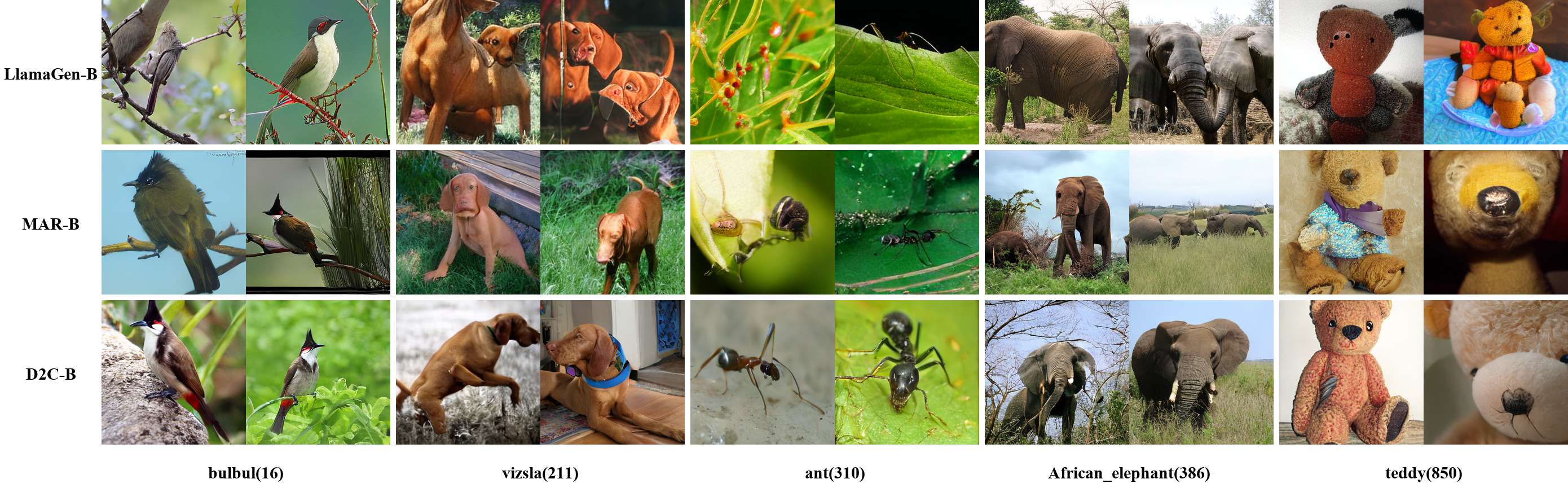}}
\caption{Visualization of samples across various classes and models.}
\label{visualization}
\end{center}
\vskip -0.1in
\end{figure*}

\section{Experiments}
\subsection{Implementation Details}
\textbf{Image Tokenizer.} We use the VQGAN \cite{esser2021taming} model from LlamaGen \cite{sun2024autoregressive} as our discrete image tokenizer, which tokenizes a $256\times256$ image into $256$ discrete-valued tokens with a $16384$ codebook size. And the VAE model \cite{rombach2022high} from LDM \cite{rombach2022high} is employed as our continuous image tokenizer with a downsampling factor $16$.

\textbf{Dataset.} Our models are trained on ImageNet \cite{deng2009imagenet} which contains $1000$ object classes of images, a total of 1, 281, 167 training images. To reduce training cost, we pretokenize each image from training set into discrete-valued tokens and continuous-valued tokens with VQGAN and VAE, respectively. Before this, each image is preprocessed with center cropping and horizontal flipping augmentation.

\textbf{Classifier-free Guidance.} The classifier-free guidance \cite{ho2022classifier} method has been demonstrated to effectively improve the generative image quality in conditional image generation tasks. During the training phase, we drop out both the class tokens and discrete-valued tokens with a probability 0.1 that are replaced with the learnable and fake class token and discrete-valued tokens, respectively. And in inference, the predicted noise follows the equation: $\varepsilon=\varepsilon_{\theta}(x_{t}|t, z_{uncond})+\omega \cdot(\varepsilon_{\theta}(x_{t}|t, z_{cond})-\varepsilon_{\theta}(x_{t}|t, z_{uncond}))$, where $\omega$ is the guidance scale, and $t$ is the timestep, $z_{cond}$ is the discrete and continuous-valued tokens condition and $z_{uncond}$ is the fake tokens.

\textbf{Training.} For convenience, we initialize our discrete-valued autoregressive model using the pretrained LlamaGen-B model \cite{sun2024autoregressive} and then freeze all parameters. Our hybrid autoregressive models of the second stage, D2C-B and D2C-L are trained for 800 epochs using the AdamW optimizer with $\beta_1=0.9$, $\beta_2=0.95$ and weight decay as $0.02$. The base learning rate is set 5e-5 per 256 batch size and the learning rate is linearly increased with the training step at first 100 epochs, then it remains constant. Moreover, we maintain the exponential moving average (EMA) of the model parameters with a momentum of 0.9999. In training, we randomly sample a mask rate from a truncated Gaussian distribution [0.7, 1.0]. The discrete token embedding layer of hybrid autoregressive model is initialized by the discrete-valued autoregressive model and freezed.

\textbf{Inference and Evaluation.} We generate each image with a simpler linear guidance schedule and progressively reduce the masking ratio from 1.0 to 0 with a cosine schedule. The discrete-valued autoregressive model produces the discrete-valued tokens with multinomial sampling strategy only given the class. In addition, we use Fr\'{e}chet Inception Distance (FID) \cite{heusel2017gans} as the primary evaluation metric, while Inception Score (IS) \cite{salimans2016improved}, as auxiliary evaluation metric.

\subsection{Main Results}
In Table.\ref{table_1}, we compare our model with popular image
generation models, include discrete-valued models \cite{esser2021taming, rombach2022high, chang2022maskgit, yu2023language, yu2024image, weber2024maskbit, sun2024autoregressive, yu2024randomized}, continuous-valued models \cite{rombach2022high, peebles2023scalable, tschannen2023givt, li2024autoregressive} and both-based models \cite{tang2024hart}. Compared with them, our models achieve superior performance with classifier-free guidance. Specifically, D2C-B with Q-Former Module achieves an FID score 2.09 only using 64 steps, significantly outperforming the LlamaGen-B \cite{sun2024autoregressive} and MAR-B \cite{li2024autoregressive} models. And it achieves $3.5\times$ higher speed than MAR in inference. When using 256 steps, D2C-B with Q-Former Module further improve the FID score to 2.06. In addition, we also get better results than the HART model of the same type \cite{tang2024hart}. Besides, we also explore the bigger model size, D2C-L with Q-Former Module getting an FID score 1.71, where the consistent performance improvement is observed when scaling up model size. These results confirm that hybrid condition merging discrete and continuous token contains more abundant information compared with the lossless continuous-valued tokens. On the other hand, we notice that the performance of our models without classifier-free guidance is not satisfactory. Because the discrete-valued tokens are sampled with classifier-free guidance when not using classifier-free guidance to generate continuous-valued tokens. This reduces diversity of generated image but significantly improves the IS score.

\textbf{Q-Former vs Cross-Attention.} We conduct experiments with diffusion fusion modules: cross-attention module and q-former module. Obviously, our model with q-former module gets better results compared with cross-attention module. This indicates that the query tokens in the q-former module are able to refine efficient information and discard useless information from discrete image features.

\begin{figure*}[th]
\begin{center}
\centerline{\includegraphics[width=\textwidth]{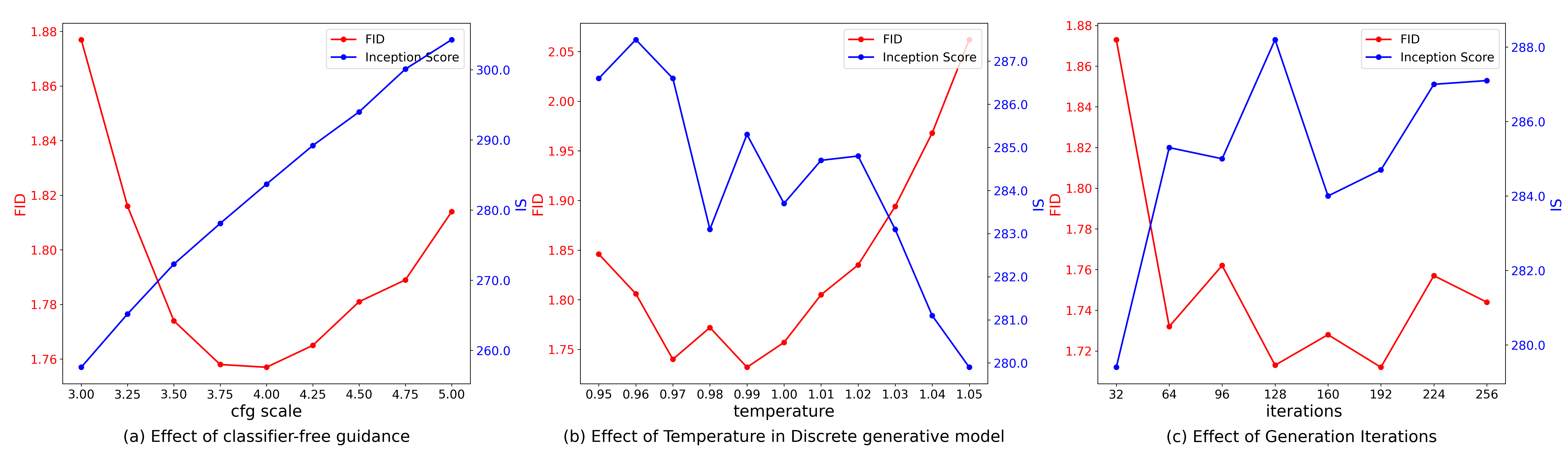}}
\caption{The variation of FID and IS with respect to different hyper-parameter: cfg scale, temperature of discrete-valued generative model and number of generation
iterations.}
\label{cfg_change_figure}
\end{center}
\vskip -0.1in
\end{figure*}

\textbf{Visualization.} As shown in Figure.\ref{visualization}, we visualize generated samples by our model, MAR \cite{li2024autoregressive} and LlamaGen \cite{sun2024autoregressive}, which demonstrates that our model is capable of generating high-quality and diverse images.

\subsection{Ablation Studies}
\textbf{Number of Query Tokens and Q-former Architecture.} As shown in Table.\ref{table_2}, we conduct ablation experiments on assess the impact about number of query tokens on model performance. In the MAR model \cite{li2024autoregressive}, 64 identical class tokens are padded at the start of the encoder sequence. In our model, a part of class tokens are replaced with query tokens extracting feature from discrete-valued tokens. Specifically, we examine the effect of $16$, $32$ and $64$ query tokens, respectively. In order to balance efficiency and performance, we chose $16$ query tokens in the q-former module. Besides, we compare different q-former architecture in Table.\ref{table_3}. It is obvious that the q-former with encoder-decoder architecture gets optimal performance. It also shows that aggregating reliable information is indispensable before extracting the important information from the discrete-valued tokens.

\begin{table}[t]
\caption{Comparison of different query tokens in q-former refiner on class-conditional Generation on ImageNet $256\times256$ benchmark, where each model with 3 MLP is trained to 400 epochs.}
\label{table_2}
\vskip 0.1in
\begin{center}
\setlength{\tabcolsep}{2.5pt}
\renewcommand{\arraystretch}{1.25}
\begin{tabular}{c|cc|cc}
\toprule
 &  \multicolumn{2}{c|}{with CFG} & \multicolumn{2}{c}{without CFG} \\
\textbf{Number of query tokens} & \textbf{FID\textdownarrow} & \textbf{IS\textuparrow} & \textbf{FID\textdownarrow} & \textbf{IS\textuparrow} \\
\midrule
16 & 2.43 & 263.8 & 3.70 & 227.7 \\
32 & 2.41 & 258.8 & 3.71 & 228.0 \\
64 & 11.70 & 81.1 & 7.34 & 115.1 \\
\bottomrule
\end{tabular}
\end{center}
\vskip -0.1in
\end{table}

\begin{table}[t]
\caption{Comparison of different architecture of q-former refiner on class-conditional Generation on ImageNet $256\times256$ benchmark, where each model with 3 MLP is trained to 400 epochs and the number of query tokens is 16.}
\label{table_3}
\vskip 0.1in
\begin{center}
\setlength{\tabcolsep}{2.5pt}
\renewcommand{\arraystretch}{1.25}
\begin{tabular}{c|cc|cc}
\toprule
 &  \multicolumn{2}{c|}{with CFG} & \multicolumn{2}{c}{without CFG} \\
\textbf{Q-former Architecture} & \textbf{FID\textdownarrow} & \textbf{IS\textuparrow} & \textbf{FID\textdownarrow} & \textbf{IS\textuparrow} \\
\midrule
Decoder & 2.55 & 270.3 & 3.81 & 223.1 \\
Encoder-Decoder & 2.43 & 263.8 & 3.70 & 227.7 \\
\bottomrule
\end{tabular}
\end{center}
\vskip -0.1in
\end{table}

\textbf{Effect of Classifier-free Guidance and Temperature of Generating Discrete Token.} In Figure.\ref{cfg_change_figure} (a), we conduct ablation experiments on different classifier-free guidance scale. Increasing the classifier-free guidance scale can improve the quality of generated images but reduce the variety, which balances the diversity and fidelity of images. On the other hand, the temperature in sampling the discrete token has a similar effect with classifier-free guidance scale in Figure.\ref{cfg_change_figure} (b). In our model with q-former module, the optimal performance is achieved while the classifier-free guidance scale is setted to 4.0. And temperature of discrete-valued generation model to 0.99 is the best choice with classifier-free guidance.

\textbf{Effect of Generation Iterations.} Figure.\ref{cfg_change_figure} (c) show FID and IS metrics variations with the number of generation Iterations. With the increase of the number of generation Iterations, both FID and IS metrics show a trend of increasing first, then slight fluctuation. Our model achieves the best result at 192 generation Iterations. In fact, using 64 steps in inference is sufficient to achieve a strong generation quality.

\section{Conclusion}
In this paper, we propose D2C, a two-stage hybrid autoregressive model, which merges the discrete-valued tokens from a small discrete-valued image generation model with the continuous-valued tokens for enhancing image synthesis in the class-to-image task. In order to connect both, we design two kinds of fusion modules, cross-attention module and q-former module, to achieve fine-grained continuous-valued tokens generation based on coarse-grained discrete-valued tokens. Our experiment results indicate that our model surpasses both discrete-valued and continuous-valued model in generation image quality. Moreover, the effectiveness of fusing discrete and continuous-valued tokens in autoregressive models is proved. We hope that this work contributes to advancing hybrid model combining continuous-valued and discrete-valued for image generation.

\section*{Impact Statement}
This paper presents work whose goal is to improve the image generation quality through fusing discrete and continuous-valued tokens in the class-to-image task. There are many potential societal consequences of our work, none which we feel must be specifically highlighted here.

% In the unusual situation where you want a paper to appear in the
% references without citing it in the main text, use \nocite
\nocite{}

\bibliography{example_paper}
\bibliographystyle{icml2025}

%%%%%%%%%%%%%%%%%%%%%%%%%%%%%%%%%%%%%%%%%%%%%%%%%%%%%%%%%%%%%%%%%%%%%%%%%%%%%%%
%%%%%%%%%%%%%%%%%%%%%%%%%%%%%%%%%%%%%%%%%%%%%%%%%%%%%%%%%%%%%%%%%%%%%%%%%%%%%%%
% APPENDIX
%%%%%%%%%%%%%%%%%%%%%%%%%%%%%%%%%%%%%%%%%%%%%%%%%%%%%%%%%%%%%%%%%%%%%%%%%%%%%%%
%%%%%%%%%%%%%%%%%%%%%%%%%%%%%%%%%%%%%%%%%%%%%%%%%%%%%%%%%%%%%%%%%%%%%%%%%%%%%%%
% \newpage
% \appendix
% \onecolumn
% \section{You \emph{can} have an appendix here.}

% You can have as much text here as you want. The main body must be at most $8$ pages long.
% For the final version, one more page can be added.
% If you want, you can use an appendix like this one.  

% The $\mathtt{\backslash onecolumn}$ command above can be kept in place if you prefer a one-column appendix, or can be removed if you prefer a two-column appendix.  Apart from this possible change, the style (font size, spacing, margins, page numbering, etc.) should be kept the same as the main body.
%%%%%%%%%%%%%%%%%%%%%%%%%%%%%%%%%%%%%%%%%%%%%%%%%%%%%%%%%%%%%%%%%%%%%%%%%%%%%%%
%%%%%%%%%%%%%%%%%%%%%%%%%%%%%%%%%%%%%%%%%%%%%%%%%%%%%%%%%%%%%%%%%%%%%%%%%%%%%%%

\end{document}